\documentclass[letterpaper]{article} % DO NOT CHANGE THIS
\usepackage{aaai24}  % DO NOT CHANGE THIS
\usepackage{times}  % DO NOT CHANGE THIS
\usepackage{helvet}  % DO NOT CHANGE THIS
\usepackage{courier}  % DO NOT CHANGE THIS
\usepackage[hyphens]{url}  % DO NOT CHANGE THIS
\usepackage{graphicx} % DO NOT CHANGE THIS
\usepackage{natbib}  % DO NOT CHANGE THIS AND DO NOT ADD ANY OPTIONS TO IT
\usepackage{caption} % DO NOT CHANGE THIS AND DO NOT ADD ANY OPTIONS TO IT
\usepackage{url}            % simple URL typesetting
\usepackage{booktabs}       % professional-quality tables
\usepackage{amsfonts}       % blackboard math symbols
\usepackage{nicefrac}       % compact symbols for 1/2, etc.
\usepackage{microtype}      % microtypography

\usepackage{graphicx}

\usepackage{amsmath}
\usepackage{amsthm}
\usepackage{algorithm}
\usepackage{algorithmic}
\usepackage{amssymb}
\usepackage{svg}
\usepackage{subfigure} 
\usepackage{MnSymbol}
\usepackage{multirow}

\usepackage{graphicx}
\usepackage{enumitem}
\usepackage{subfigure}
\usepackage{amssymb}
\usepackage{cancel}
\usepackage{caption}
\usepackage{array}

\usepackage{amssymb}
\usepackage{tabularx}
\usepackage{bm}

\usepackage{newfloat}
\usepackage{listings}
\DeclareCaptionStyle{ruled}{labelfont=normalfont,labelsep=colon,strut=off} % DO NOT CHANGE THIS
\floatstyle{ruled}
\newfloat{listing}{tb}{lst}{}
\floatname{listing}{Listing}
\title{Graph Contrastive Invariant Learning from the Causal Perspective}
\author{
    Yanhu Mo\textsuperscript{\rm 1},
    Xiao Wang\textsuperscript{\rm 2}$^*$,
    Shaohua Fan\textsuperscript{\rm 3,4},
    Chuan Shi\textsuperscript{\rm 1}\thanks{Corresponding Author.}
}

\affiliations{
    \textsuperscript{\rm 1}Beijing University of Posts and Telecommunications\\
    \textsuperscript{\rm 2} Beihang University \\
    \textsuperscript{\rm 3} Tsinghua University \\
    \textsuperscript{\rm 4} Key Laboratory of Big Data \& Artificial Intelligence in Transportation, Ministry of Education(Beijing Jiaotong University) \\
    \{moyanhu, shichuan\}@bupt.edu.cn, xiao\_wang@buaa.edu.cn, fanshaohua@bupt.cn
}

\usepackage{bibentry}
\begin{document}

\maketitle

\begin{abstract}
Graph contrastive learning (GCL), learning the node representation by contrasting two augmented graphs in a self-supervised way, has attracted considerable attention. GCL is usually believed to learn the invariant representation. However, \textit{does this understanding always hold in practice?} In this paper, we first study GCL from the perspective of causality. By analyzing GCL with the structural causal model (SCM), we discover that traditional GCL may not well learn the invariant representations due to the non-causal information contained in the graph. \textit{How can we fix it and encourage the current GCL to learn better invariant representations?} The SCM offers two requirements and motives us to propose a novel GCL method. Particularly, we introduce the spectral graph augmentation to simulate the intervention upon non-causal factors. Then we design the invariance objective and independence objective to better capture the causal factors. Specifically, (i) the invariance objective encourages the encoder to capture the invariant information contained in causal variables, and (ii) the independence objective aims to reduce the influence of confounders on the causal variables. Experimental results demonstrate the effectiveness of our approach on node classification tasks.
\end{abstract}

\section{Introduction}

Graph Neural Networks (GNNs) learn node representations by aggregating information from neighborhoods, which have received a great deal of attention and achieved competitive performance on various tasks over the past few years \cite{kipf2016semi,velivckovic2017graph,hamilton2017inductive}. Despite the great success, most GNNs are trained with node labels, while it is well known that manual annotations are expensive and hard to collect; therefore, self-supervised learning (SSL) has gained popularity due to its competitive performance and label-free setting \cite{chen2020simple,he2020momentum}. Graph Contrastive Learning (GCL), as one of the most successful strategies for self-supervised representation learning on graphs, has shown state-of-the-art performance on many downstream tasks \cite{velivckovic2018deep,zhu2020deep,qiu2020gcc, chen2023universal}.

The typical GCL method mainly includes three parts: graph augmentation, encoding architecture, and contrastive loss. Most existing GCL methods learn the representations by comparing the augmentations. First, GCL generates two augmented graphs from the original graph based on some augmentation strategies, e.g., dropping edges \cite{velivckovic2018deep}. Then the two augmented graphs are fed into the encoding architecture (e.g., GCN \cite{kipf2016semi}) to learn the node representations. Finally, the contrastive loss (e.g., InfoNCE \cite{zhu2020deep}) is used to train the GCL model by making the representations of positive pair in two augmented graphs similar and representations of negative pairs dissimilar. It is believed that GCL is able to learn the invariant representations by contrasting the positive and negative pairs \cite{zhu2021graph,liu2022revisiting}. The learned invariant representations will be beneficial for the downstream tasks.

Generically, the question we want to ask is: \textit{does GCL always possess the invariant representation ability in practice? When will it fail and how to enhance this ability?} A well-informed answer can provide a deeper understanding of the learning mechanism of GCL, identify the weakness of GCL, and motivate more powerful GCL to be proposed. The invariant representation usually represents the essential information, which can also be considered as some kind of causal variables in a graph \cite{arjovsky2019invariant,wu2022discovering,fan2023generalizing}. This natural and intuitive connection inspires us to explore the GCL from the causal perspective. We start with a causal analysis of GCL (more details are in Section \ref{section:3}) based on the structural causal model (SCM). The SCM indicates that if there are both causal and non-causal factors in a graph, only when the causal factors are the same and the non-causal factors are different in the original and two augmented graphs, GCL is able to learn the invariant causal factors. However, with the graph structures change by graph augmentation strategies, e.g., random augmentation, it is very hard to guarantee that and finally weakens the invariant representation learning ability.

Once the weakness is identified, a natural question is \textit{how can we fix it and improve the invariant learning ability?} Still, SCM offers the following two requirements for GCL: one is on the augmentation mechanism, i.e., a better graph augmentation strategy should take the causal and non-causal factors into account. Without distinguishing the two factors, the representations obtained by GCL methods may contain both causal and non-causal information, which might weaken the performance on downstream tasks. The second is on the learning mechanism, i.e., given causal and non-causal factors, how to ensure that GCL is able to learn the causal variables. This is extremely challenging because there is usually no prior knowledge of labels, etc. Moreover, even for different causal variables, they may be also statistically dependent due to the backdoor path caused by the confounders.

In this paper, we propose a novel graph contrastive learning method from the standpoint of causality, called GCIL\footnote{https://github.com/BUPT-GAMMA/GCIL} (\textbf{G}raph \textbf{C}ontrastive \textbf{I}nvariant \textbf{L}earning). Specifically, we first elaborate the structural causal model (SCM) to describe the graph generation process. According to the SCM, we generate two views from the perspective of the graph spectrum, which simulate perturbing the non-causal factors while keeping the causal contents unchanged. We assume each dimension in the representations holds the Gaussian distribution, and then we propose an invariance objective to guarantee that the representations generated from two views maintain the same mean and standard deviation for each dimension. Therefore, we can extract the causal information from the original graph. Finally, we utilize the independent module to push the causal factors mutually independent, thus none of them influence others by the backdoor path. Our contributions are summarized as follows:
\begin{itemize}
	\item We study graph contrastive learning from the perspective of causality, and point out that existing methods may not learn the invariant representations due to the non-causal information contained in the graph.
	\item Based on the theory of causality, we propose a novel graph contrastive learning method that aims to learn an invariant representation via extracting the causal information contained in the original graph.
        \item We validate the effectiveness of GCIL compared with state-of-the-art methods on four datasets, and our method outperforms both semi-supervised and self-supervised baselines.
\end{itemize}

\begin{figure}[h]
\centering
\includegraphics[scale=0.8]{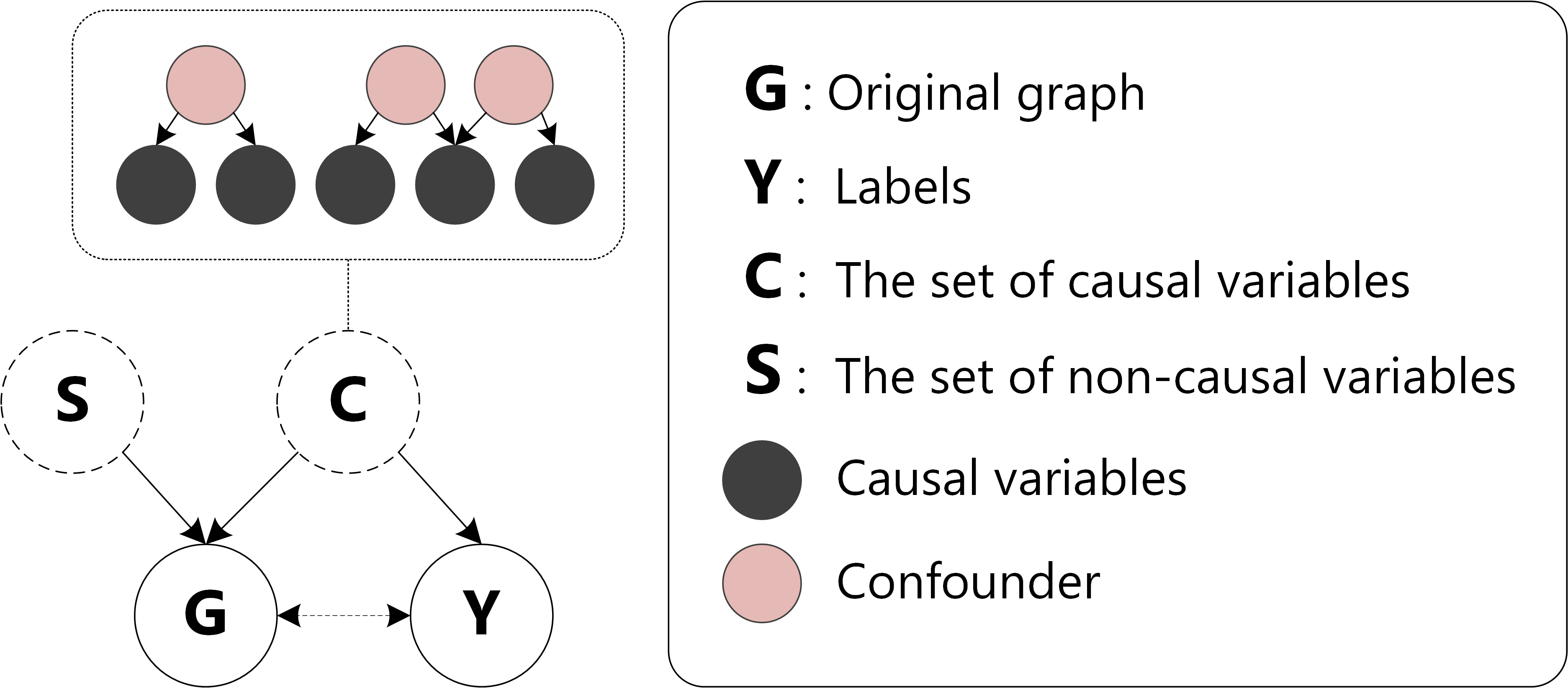}
\caption{SCM of the graph generation process. The dashed circle and solid circle represent unobserved and observed variables, respectively. }
\label{scm}
\end{figure}

\section{Related Work}
\textbf{Graph Neural Networks.} Graph Neural Networks (GNNs) show outstanding performance in various tasks recently. For example, GCN \cite{kipf2016semi} averages the information of one-hop neighbors. GAT \cite{velivckovic2017graph} assigns different weights to different neighbors. GraphSAGE \cite{hamilton2017inductive} aggregates a subset of neighbors using various pooling methods. In the \citet{bruna2013spectral}, fourier bias is used to decompose graph signals; ChebNet \cite{defferrard2016convolutional} improves efficiency with the Chebyshev expansion of the graph Laplacian. Recent surveys \cite{wu2020comprehensive,zhou2020graph} provide a more complete review.

\noindent \textbf{Graph Contrastive Learning.}
Self-supervised representation learning has attracted considerable attention in computer vision \cite{oord2018representation,chen2020simple,he2020momentum, wang2017community}. Motivated by the local-global mutual information maximization, DGI \cite{velivckovic2018deep} contrasts local node embedding and global summary vector. MVGRL \cite{hassani2020contrastive} employs diffusion or distance matrices to handle contrastive tasks. GRACE \cite{zhu2020deep}, GraphCL \cite{you2020graph}, GCA \cite{zhu2021graph} obtain two views by perturbing the original graph, and then they learn representations using InfoNCE loss. ProGCL \cite{xia2022progcl} use of EM algorithm to samples more appropriate hard negative samples to learn node embeddings. CCA-SSG \cite{zhang2021canonical} optimizes a feature-level objective rather than discriminating positive and negative samples. Recent advances in graph contrastive learning have been summarized in several surveys \cite{liu2022graph,xie2022self}.

\noindent \textbf{Causality in Graph Neural Networks.} Causality studies the relationship between variables\cite{pearl2016causal,pearl2009causality}, which has shown many benefits in deep learning. With the help of causality, many methods have achieved great success in various computer vision tasks \cite{zhang2020causal,mitrovic2020representation}. There is some research related to the graph. For example, DIR-GNN \cite{wu2022discovering} conducts the intervention on the non-causal part to generate the representations to discover the rationale on the graph. DisC \cite{fan2022debiasing} disentangles the graph into causal and bias subgraphs, thus alleviating the bias in the datasets. RGCL \cite{li2022let} introduces an invariance look into self-supervised learning and proposes a method to preserve stable semantic information. \citet{fan2022debiased} explores agnostic label selection bias in GNNs. CIGA \cite{chen2022learning} guarantees OOD generalization across distribution shifts by capturing the invariance of graphs. GraphNOTEARS \cite{fan2023directed} studies the associated mechanism for generating node features in dynamic graph data. Different from them, we exploit the self-supervised node classification task from the standpoint of causality and propose a new graph contrastive learning loss based on the theory of causality.

\section{Causal Analysis on GCL}
\label{section:3}
\subsection{Notations and Framework}
\textbf{Notations.} Let $G = (V, E)$ denote a graph, where $V$ is the set of nodes with $|V| = N$ and $E$ is the set of edges. Each graph $G$ has an adjacency matrix $A^{N\times N}$, where $A_{ij}$ denotes the relation between two nodes, i.e., $A_{ij} = 1$ iff there is an edge between $v_i$ and $v_j$. Graph $G$ is often assigned with a node feature matrix $X=[x_1,x_2,...,x_N]^T\in R^{N\times F}$, where $x_i$ is the feature vector of node $v_i$. The goal of GCL is to learn an optimal GNN encoder to acquire node representations without requiring any label information.

% GCN \cite{kipf2016semi} is usually chosen as the encoder, which aggregates neighbors' information iteratively. Formally, the $i^{th}$ GCN layer can be written as:
% \begin{equation}
%     G C N\left(\boldsymbol{A}, \boldsymbol{H}^{(i)}\right)=\boldsymbol{D}^{-1 / 2} \boldsymbol{A} \boldsymbol{D}^{-1 / 2} \boldsymbol{H}^{(i-1)} \boldsymbol{W}^i,
% \end{equation}
% where $D$ denotes the degree matrix of $A$, and $H^i$ denotes the node embeddings in the $i^{th}$ layer, and ${H^{(0)} = X$. $W^i$ represents the weight matrix.

\noindent \textbf{Graph Contrastive Learning (GCL).} Given graph $G$, the pipeline of GCL is to generate two graph views $V_A, V_B$ with graph augmentation generators $\tau_A, \tau_B$ as follows:
\begin{equation}
    V_A=\tau_A(G,X), V_B=\tau_B(G,X).
\end{equation}
 Then, the two graph views are fed into a shared GNN encoder $g(\cdot)$ to generate the node representations $Z^A, Z^B$, respectively:
 \begin{equation}
    Z^A=g(V_A), Z^B=g(V_B),
    \label{get representations}
\end{equation}
where $Z^A=[z^A_1,z^A_2,...,z^A_N]^T\in R^{N\times d}$ and $Z^B=[z^B_1,z^B_2,...,z^B_N]^T\in R^{N\times d}$. The $d$ denotes the dimension of representation. We employ a contrastive loss \cite{velivckovic2018deep,chen2020simple} to optimize the encoder $g(\cdot)$, pushing it to construct the invariant representations. Once the encoder $g(\cdot)$ is well trained, we can finally obtain the node representations $Z=g(G, X)$.

\subsection{Causal Interpretation}
\label{section:3.2}
Before analyzing graph contrastive learning with a causal view, we first elaborate on the structural causal model (SCM) to describe the graph generation process based on the following assumptions: (1) The original graph $G$ can be decomposed into the set of causal variables $C$ and the set of non-causal variables $S$. (2) Only $C$ causally influences both the input graph $G$ and downstream tasks, and $S$ does not provide any information about the downstream tasks. (3) There is no causal relationship between $S$ and $C$., i.e., the generation process of $C$ is independent of $S$. Based on these assumptions, the SCM for the node classification task can be depicted in Figure \ref{scm}, where $Y$ represents the node labels. The dashed circle in SCM means the unobserved latent variable and the solid circle represents the observed variable. Please note that the causal variables in $C$ may also be statistically dependent, so when we estimate the causal effect of a causal variable (marked by gray nodes), other causal variables are considered confounders (marked by red nodes). More details between variables are described as follows:

\begin{itemize}
	\item $C \rightarrow G \leftarrow S$. The observed node data is generated by two unobserved latent variables: causal variables $C$ and non-causal variables $S$.
	\item $C \longrightarrow Y$. This link means the causal part $C$ contains all the necessary information for the node classification task.
        \item $G \dashleftarrow \dashrightarrow Y$. This dashed link indicates statistical dependence between $G$ and $Y$. Since the raw graph contains two kinds of information, we will construct the statistical dependence relationship between $G$ and $Y$ if we do not distinguish the latent variables $C$ and $S$.
        \item Common causes between variables are called confounders. In the SCM, the confounder provides a backdoor path to the causal variables, contributing to the correlation between different causal variables.
        
\end{itemize}

Now we analyze GCL based on the above SCM. In GCL, the augmented graphs $V_A$ will contain the causal variables $C_A$ and non-causal variables $S_A$, also, $V_B$ will contain $C_B$ and $S_B$. Ideally, only when $C_A=C_B=C$ and $S_A \neq S_B$, GCL is able to capture the invariant information $C$ well. However, let us take the widely used graph augmentation strategy (i.e., random augmentation \cite{zhu2020deep}) as an example. Because the random augmentation does not distinguish $C$ and $S$, it is very hard to guarantee that $C_A=C_B=C$ and $S_A \neq S_B$ still hold after graph augmentation. As a result, the learned final representation $Z$ will contain both $C$ and $S$, and the predicted label $Y'$ will be made as follows: 
\begin{equation}
    h(Z) = h(C,S) = Y',
\end{equation}
where $h$ denotes a classification head. Apparently, because the representation contains non-causal information, the prediction might change when the non-causal part shifts, i.e.:
\begin{equation}
    h(C,S) \neq h(C,S').
\end{equation}
It is unreasonable because the two representations contain the same causal information, yet their predictions differ.

Therefore, we should perturb the non-causal factors $S$ while keeping the causal factors $C$ unchanged when generating augmentations, which can be regarded as conducting intervention on $S$, and meanwhile, ensure the consistency of causal information between different augmentations. Thus, the model should satisfy the following equation:
\begin{equation}
    P^{d o\left(S=s_i\right)}\left(Y | C\right)=P^{d o\left(S=s_j\right)}\left(Y | C\right),
    \label{intervention}
\end{equation}
where $d o\left(S=s\right)$ represents the intervention on non-causal factors $S$. This formulation encourages the model to extract information only contained in causal factors $C$ and discard trivial information. 

Moreover, as shown in Figure \ref{scm}, the causal variables contained in the $C$ might be influenced by confounders (i.e., $c_1 \leftarrow Confounder \rightarrow c_2$, where $c_1,c_2$ denote two causal variables, respectively). The confounders lead to the correlation between variables, indicating that different causal variables will encode similar information. To obtain a more informative representation, it is essential to eliminate the influence of the confounders, which refers to the mutually independent between causal variables, i.e.,
\begin{equation}
    \forall c_i,c_j \in C, c_i \upmodels c_j,
    \label{variables independence}
\end{equation}
where $c_i$ and $c_j$ denote two different causal variables. This formulation implies that different variables in $C$ are mutually independent, effectively mitigating the influence of confounders.

\begin{figure*}[t]
\centering
\includegraphics[width=0.8\textwidth]{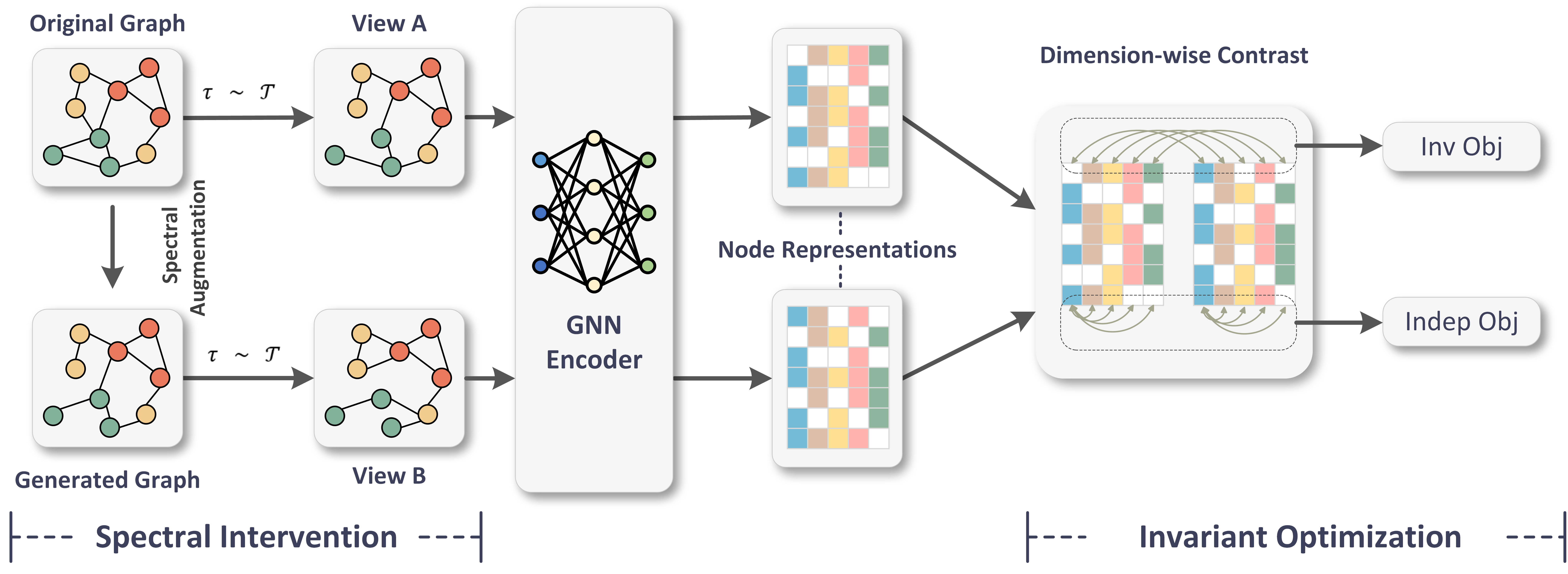}
\caption{Overview of the GCIL framework. Given an original graph G, we first generate two views by spectral and random augmentation. The two views are subsequently fed into a shared GNN encoder to generate representations. At last, we optimize the invariance objective and the independence objective to render the model to learn the invariant representations.}
\label{framework}
\end{figure*}

\section{The Proposed Model: GCIL}
In this section, we illustrate our proposed graph contrastive learning algorithm inspired by causality. The overall framework is shown in Figure \ref{framework}. We initially generate two views through spectral graph augmentation and random augmentation, which can be seen as causal interventions. The two views are subsequently fed into a shared GNN encoder to generate node representations. Finally, we propose two dimension-level contrastive objectives. The invariance objective pushes the encoder to capture the invariant information contained in causal variables, while the independence objective encourages the different dimensions of the representations to be independent. Thus, we can obtain the invariant representations from the well-trained model.

\subsection{Causal Intervention}
Based on the above analysis, we need to generate graph augmentations that meet the following conditions: perturbing the non-causal information and maintaining causal information. However, it is very challenging that graph $G$ does not provide any prior knowledge about $C$ and $S$. Following \cite{liu2022revisiting}, perturbing the structure of the initial graph will change the strength of frequencies in the graph spectrum. The general graph augmentation should satisfy that: in two contrasted augmentations, the difference in high-frequency amplitudes should be greater than in low-frequency amplitudes. In other words, the lowest-frequency information can be approximately regarded as an invariant pattern between the two views. 

We consider the low-frequency information in the graph as causal content and the high-frequency information as non-causal content. Therefore, we conduct the intervention on $S$ by disrupting the high-frequency information while leaving the low-frequency information unchanged. In particular, given an original adjacency matrix $A$, our goal is to obtain an adjacency matrix $A'$ by intervention, i.e., $A' = A + \Delta_{\boldsymbol{A}+} - \Delta_{\boldsymbol{A}-}$, where $\Delta_{\boldsymbol{A}+}$ and $\Delta_{\boldsymbol{A}-}$ indicate which edge is added and deleted, respectively. $\Delta_{\boldsymbol{A}+}$ is obtained by maximizing the following objective function:
\begin{equation}
\begin{aligned}
    J &= \left\langle \Theta L, \Delta_{A+} \right\rangle^2 + \epsilon H\left(\Delta_{A+}\right) \\
    &\quad + \left\langle f, \Delta_{A+} \mathbf{1}_n - a \right\rangle + \left\langle g, \Delta_{A+}^\top \mathbf{1}_n - b \right\rangle,
\end{aligned}
\end{equation}

where $\Theta$ is a parameter updated in training and $L$ is the laplacian matrix of $G$. $\epsilon$ is the weight parameter, $f$ and $g$ are Lagrange multipliers, and $a$, $b$ are the node degree distributions. $H(\cdot)$ represents the entropy regularization, i.e., $H(P)=-\sum_{i, j} P_{i, j}(\log (P_{i, j})-1)$. The calculation of $\Delta_{A-}$ can be referred to \cite{liu2022revisiting}. We then corrupt two graphs with random data augmentation as follows:

\begin{equation}
    V_A = \tau_A(A,X), V_B = \tau_B(A',X), \tau_A, \tau_B \in T,
    \label{two views}
\end{equation}
where $T$ represents the whole augmentation function space.

\subsection{Invariance Objective}
Now, based on the analysis in Section \ref{section:3.2}, we need to require that the relationship between $V_A$ and $V_B$ in Eq. \ref{two views} satisfies Eq. \ref{intervention}, so that the model can learn invariant representation. 
However, GCL does not provide any information about labels $Y$. To achieve the goal in Eq. \ref{intervention}, the equation can be reformulated as follows:
\begin{equation}
    CE(C, S=s_i)=CE(C, S=s_j),
\end{equation}
where $CE$ represents the causal effect of variables. That is, we need to capture the consistency between the node representations  $Z^A, Z^B\in R^{N\times d}$ obtained by Eq. \ref{get representations}. Then, we assume that each dimension in the representation holds the Gaussian distribution, and we propose an invariance objective that encourages the representations to remain unchanged dimension-wise, i.e., by enforcing the consistency of the mean and standard deviations for each dimension of the representations. Formally, the learning objectives can be formulated as follows:
\begin{equation}
    \min_{{g}} \sum_i\left\|Z_i^A-Z_i^B\right\|_2^2,\quad s.t. Std(Z_i^A)=Std(Z_i^B)=\lambda,
    \label{inv_loss}
\end{equation}
where $Z^A_i$, $Z^B_i$ denote the $i$-th dimension of two embedding matrices, respectively, and $Std$ represents the standard deviation. The first term encourages the mean of two embedding matrices to be equal in the same dimension, and the second term pushes the standard deviation close to $\lambda$, where $\lambda$ is the hyper-parameter.

\subsection{Independence Objective}
According to the SCM, different causal variables might be correlated due to the confounders, leading to less informative representations generated by the model. To mitigate this issue, we propose an independence objective to satisfy Eq. \ref{variables independence}. The independence objective seeks to force the different causal variables to be mutually independent, thereby eliminating the correlation between causal variables. Specifically, we use the Hilbert-Schmidt Independence Criterion (HSIC) to measure the independence between factors. A value of 0 for HSIC indicates that the two variables are independent. Minimizing the following equation different dimensions in the representation $Z$ to be independent:
\begin{equation}
   \sum_{i \neq j}  HSIC(Z_i, Z_j) = \sum_{i \neq j}  \frac{1}{(N-1)^2} \operatorname{Tr}(\boldsymbol{K}_i \boldsymbol{H} \boldsymbol{K}_j \boldsymbol{H}),
\end{equation}
where $Z_i$ is the i-th dimension of embedding matrix, $\boldsymbol{H}$ represents the centering matrix $\boldsymbol{I}-\mathbf{1 1}^T \frac{1}{N}$, $\boldsymbol{K}_i$ and $\boldsymbol{K}_j$ are the two kernel matrices of $Z_i$, $Z_j$ respectively.

Here, we introduce the notion of a kernel matrix $\boldsymbol{K_i}\in R^{N\times N}$, which calculates the kernel function value between each sample of the $i$-th variable. For example, ${K_i}^{a,b} = \kappa(Z_{i,a},Z_{i,b})$, where $\kappa$ denotes the kernel function, ${K_i}^{a,b}$ represents the number in the $a$-th row and $b$-th column of the matrix $K_i$, and $Z_{i,a}$ denotes the value in the $i$-th column and $a$-th row in the representation matrix $Z$. Using a complex kernel (e.g., Gaussian kernel) in HSIC to measure independence between dimensions can result in high space complexity, making it challenging to implement in scenarios with large sample sizes and dimensions. Inspired by lemma 1 in \cite{mialon2022variance}, minimizing the HSIC of different dimensions can be considered as minimizing the sum of the off-diagonal elements of the covariance matrix. The proof for this assertion is as follows:

% Let kernel function $\kappa(Z_{i,a},Z_{i,b})=g(Z_{i,a}) g(Z_{i,b})^T$, where $g$ : $R \longrightarrow R^L$ is an elementwise projector. We denote the mapping of such projectors on $Z$ as $Q = g(Z) = [g(Z_1), . . . , g(Z_d)] \in \mathbb{R}^{N \times D L}$. According to the lemma 1 in \citet{mialon2022variance}, we have:

Let kernel function $\kappa(Z_{i,a},Z_{i,b})=g(Z_{i,a}) g(Z_{i,b})^T$, where $g : R \rightarrow R^L$ is an elementwise projector. We denote the mapping of such projectors on $Z$ as $Q = g(Z) = [g(Z_1), \ldots, g(Z_d)] \in R^{N \times D L}$. According to the lemma 1 in \citet{mialon2022variance}, we have:

\begin{equation}
    \begin{split}
    &\quad HSIC(Z_i, Z_j)\\
    &=\frac{1}{(N-1)^2} \operatorname{Tr}(g(Z_i) g(Z_i)^T \boldsymbol{H} g(Z_j) g(Z_j)^T \boldsymbol{H}) \\
    &=\frac{1}{(N-1)^2}\left\|g(Z_i)^T \boldsymbol{H} g(Z_j)\right\|_F^2 \\
    &=\| \operatorname{Cov}(g(Z_i), g(Z_j) )\|_F^2 \\
    &=\left\|\operatorname{Cov}(\boldsymbol{Q})_{(i-1)L: iL, (j-1)L: jL}\right\|_F^2.
     % \implies \sum_{i\not = j} HSIC(z^A_i,z^A_j)&=\left\|\operatorname{Cov}(\boldsymbol{Q}) \odot\left(\left(1-\boldsymbol{I}_{D}\right) \otimes \mathbf{1}_{L} \mathbf{1}_{L}^{T}\right)\right\|_{F}^{2}, 
    \end{split} 
\end{equation}

where $\operatorname{Tr}$ represents the trace of matrix, $\operatorname{Cov}$ represents the covariance of two variables. We let $g(X)=X$, i.e. linear kernel, in this case $Z = Q$, thus we have
\begin{equation}
\sum_{i \neq j} HSIC(Z_i,Z_j)=\sum_{i \neq j} \operatorname{Cov(Q)}_{i, j}^2 =\sum_{i \neq j}\operatorname{Cov(\boldsymbol{Z})}_{i, j}^2.
\label{loss_dep}
\end{equation}

The calculation of HSIC values for different dimensions is converted to the calculation of covariance. Minimizing the eq. \ref{loss_dep} ensuring independence between different dimensions.

\subsection{Optimization Objective}
We further normalize the embedding matrix dimension-wisely and $\tilde{\mathbf{Z}}$ represents the node embeddings after normalization. Note that $\left\|\Tilde{Z}_i\right\|^2$=1, thus $\min_{{g}} \sum_i\left\|Z_i^A-Z_i^B\right\|_2^2$ can be replaced by maximizing the inner product of $\Tilde{Z}_i^A$ and $\Tilde{Z}_i^B$, e.g. $\max_{{g}} \sum_{i}  \Tilde{Z}_i^A \cdot \Tilde{Z}_i^B$, where $\cdot$ represents inner product. The $s_i$ denotes the standard deviation of the $i$-th dimension before normalization. Minimizing $\sqrt{\left\|s_i-\lambda\right\|_2^2}$ pushes the standard deviation close to $\lambda$.

The overall optimization objective of our proposed GCIL is summarized as follows:

% \begin{equation}
%     \begin{split}
%         \mathcal{L} = -\alpha\sum_{i}  \Tilde{Z}_i^A \cdot \Tilde{Z}_i^B + \beta\sum_{i}\{{\sqrt{\left\|s_i^A-\lambda\right\|_2^2}} +{\sqrt{\left\|s_i^B-\lambda\right\|_2^2}}\} \\+ \gamma\sum_{i \neq j} \  \operatorname{Cov({\Tilde{Z}^A})}_{i, j}^2 + \operatorname{Cov({\Tilde{Z}^B})}_{i, j}^2,
%     \label{loss_full}
%     \end{split}
% \end{equation}
\begin{equation}
    \begin{split}
        L = -\alpha\sum_{i}  \Tilde{Z}_i^A \cdot \Tilde{Z}_i^B + \beta\sum_{i}\{\sqrt{\|s_i^A-\lambda\|_2^2} +\sqrt{\|s_i^B-\lambda\|_2^2}\} \\+ \gamma\sum_{i \neq j} \  \operatorname{Cov(\Tilde{Z}^A)}_{i, j}^2 + \operatorname{Cov(\Tilde{Z}^B)}_{i, j}^2,
    \label{loss_full}
    \end{split}
\end{equation}

where $\alpha$, $\beta$, and $\gamma$ are hyper-parameters controlling the importance of each term in the loss. The $\lambda$ represents the desired standard deviation of the dimensions.

\renewcommand\arraystretch{1.15}
\begin{table}[t]
    % \captionsetup{skip=5pt}
    \centering
    
    \begin{tabular}{lcccc} \hline
        \toprule 
        Dataset     &       \#Nodes       &       \#Edges       &  \#Classes      &       \#Features  \\ 
        \midrule
        Cora      & 2,708          & 10,556   
                  & 7              & 1,433         \\
        Citeseer  & 3,327          & 9,228       
                  & 6              & 3,703        \\
        Pubmed    & 19,717         & 88,651        
                  & 3            & 500        \\
        Wiki-CS    & 11,701         & 432,246        
                  & 10            & 300        \\
        Flickr    & 7,575         & 479,476       
                  & 9            & 12,047        \\
        \bottomrule
    \end{tabular}
    \caption{Statistics of benchmark datasets}
    \label{dataset_table}
\end{table}

\section{Experiements}

\begin{table*}[ht]
        % \footnotesize
        \renewcommand\arraystretch{1.1}
	\centering
        \setlength{\tabcolsep}{3.5pt} % 调整列间距
        % \fontsize{9.5}{12}\selectfont

	\label{tab:mainexp}

        \begin{tabular}{c|cc|cc|cc|cc|cc}

\hline
Dataset & \multicolumn{2}{c|}{Cora}             & \multicolumn{2}{c|}{Citeseer}         & \multicolumn{2}{c|}{PubMed}           & \multicolumn{2}{c|}{Wiki-CS}          & \multicolumn{2}{c}{Flickr}            \\ \hline
Metrics & Ma-F1          & Mi-F1          & Ma-F1          & Mi-F1          & Ma-F1          & Mi-F1          & Ma-F1          & Mi-F1          & Ma-F1          & Mi-F1          \\ \hline\hline
GCN     & 80.6±0.7          & 81.5±0.6          & 68.1±0.5          & 70.9±0.5          & 78.5±0.3          & 78.9±0.3          & 73.2±0.8          & 77.5±0.4          & 48.9±1.6          & 50.2±1.2          \\
GAT     & 81.3±0.3          & 82.3±0.2          & 67.5±0.2          & 72.0±0.9          & 77.4±0.2          & 77.8±0.2          & 75.5±0.4          & 78.3±0.4          & 35.0±0.8          & 37.1±0.3          \\ \hline\hline
DGI     & 80.4±0.7          & 82.0±0.5          & 67.7±0.9          & 71.7±0.8          & 76.8±0.9          & 76.7±0.9          & 70.6±0.1          & 75.6±0.1          & 31.2±1.6          & 33.0±1.6          \\
MVGRL   & 81.5±0.5          & 82.8±0.4          & 66.8±0.7          & 72.5±0.5          & 79.8±0.4          & 79.7±0.3          & 74.9±0.1          & 78.1±0.1          & 31.2±2.9          & 33.4±3.0          \\
GRACE   & 79.2±1.0          & 80.0±1.0          & 65.1±1.2          & 68.7±1.1          & 80.0±0.7          & 79.9±0.7          & 74.8±0.2          & 78.2±0.1          & 35.7±1.3          & 37.3±1.0          \\
GCA     & 79.9±1.1          & 81.1±1.0          & 62.8±1.3          & 65.9±1.0          & 80.8±0.6          & 81.4±0.6          & 74.9±0.0          & 78.3±0.0          & 41.2±0.5          & 42.2±0.6          \\
GraphCL & 80.7±0.9          & 82.3±0.9          & 67.8±1.0          & 71.9±0.9          & 77.0±0.4          & 76.8±0.5          & -                 & -                 & 32.1±1.1          & 34.5±0.9          \\
CCA-SSG & 82.9±0.8          & 83.6±0.9          & 67.9±1.0          & 73.1±0.7          & 80.7±0.6          & 81.0±0.6          & 73.7±0.2          & 78.2±0.1          & 37.0±1.1          & 39.3±0.9          \\
COSTA   & 81.2±0.4          & 82.4±0.7          & 62.9±3.6          & 66.4±4.3          & 79.8±0.6          & 80.4±0.9          &72.2±5.1           &  76.2±2.3         & \textbf{42.0±2.6} & 42.8±2.5          \\
ProGCL  & 81.0±0.4          & 81.9±0.6          & 62.9±0.9          & 65.9±0.9          & 80.5±2.3          & 80.5±2.3          & 75.3±0.1          & \textbf{78.7±0.1} & 37.8±1.1          & 38.5±1.0          \\
Ours    & \textbf{83.8±0.5} & \textbf{84.4±0.7} & \textbf{69.1±0.4} & \textbf{73.7±0.5} & \textbf{81.5±0.5} & \textbf{81.6±0.7} & \textbf{75.6±0.1} & 78.6±0.1          & 40.0±0.8          & \textbf{43.0±0.4}

\\ \hline

\end{tabular}
\caption{Quantitative results on node classification, where both the mean accuracy and the standard deviation are shown in the table. The best performance of self-supervised is bolded in table. The '-' indicates Out-of-Memory on a 24GB GPU.}
\label{Performance}
\end{table*}

\subsection{Experimental Setup}
\textbf{Dataset.} To evaluate our method, we consider five commonly used node classification benchmark datasets from the previous works \cite{velivckovic2018deep,mernyei2020wiki,liu2022revisiting}, including Cora, Citeseer, Pubmed, Wiki-CS, and Flickr. 
The statistic of these datasets is summarized in Table \ref{dataset_table}. 
In the above datasets, three benchmark citation datasets contain sparse one-hot features, while Wiki-CS has dense numerical features. We adopt the public splits for Cora, Citeseer, Pubmed, and Flickr, where the training set contains 20 nodes per class, 500 nodes for validation, and 1000 for testing. For the Wiki-CS dataset, we evaluate the models on the public splits provided in \cite{mernyei2020wiki}. More details of datasets are presented in Appendix A.

\noindent \textbf{Baseline.} 
We choose two kinds of methods as
benchmarks: semi-supervised methods and self-supervised methods. \textbf{(1) Semi-supervised mehtods}: GCN \cite{kipf2016semi} and GAT \cite{velivckovic2017graph}. \textbf{(2) Self-supervised methods}: DGI \cite{velivckovic2018deep}, MVGRL \cite{hassani2020contrastive}, GRACE \cite{zhu2020deep}, GCA \cite{zhu2021graph}, GraphCL \cite{you2020graph}, COSTA \cite{zhang2022costa}, CCA-SSG \cite{zhang2021canonical}, ProGCL\cite{xia2022progcl}.

\noindent \textbf{Parameter Settings.} We set the dimensions of all datasets to 512, with a learning rate of 0.0002 for the Flickr dataset and 0.001 for all other datasets. The weight decay for all datasets is 0.0001. Additional details of parameter settings are presented in Appendix C.

\begin{table}[t]
\centering
\renewcommand\arraystretch{1.1}

\begin{tabular}{|l|cc|cc|} 
\hline
\multirow{2}{*}{Ablation} & \multicolumn{2}{c|}{Cora} & \multicolumn{2}{c|}{Wiki-CS} \\ \cline{2-5}
& {Ma-F1} & {Mi-F1} & {Ma-F1} & {Mi-F1} \\
\hline
w/o indep & 57.5  & 57.8& 34.3 & 51.4 \\ 
w/o inv & 79.2  & 80.5& 57.5 & 63.6 \\ 
w/o Aug & 83.3  & 84.1 & 75.1& 78.0 \\ 
GCIL & \textbf{83.8}  & \textbf{84.4} & \textbf{75.6}& \textbf{78.6} \\
\hline
\end{tabular}
\caption{Ablation Results on Cora and Wiki-CS}
\label{ablation}
\end{table}

\subsection{Node Classification}
In this section, we evaluate the proposed GCIL on node classification. We conducted experiments on five datasets and the results are shown in Table \ref{Performance}. The best results of all self-supervised learning methods are bolded in the table. Please note that GraphCL has an out-of-memory issue on the Wiki-CS dataset. As we can see, our method GCIL achieves excellent performance on all five datasets compared to self-supervised methods. Except for the Flickr dataset, our method performs even better than semi-supervised learning methods. Specifically, we achieve the SOTA performance on Cora, Citeseer, and Pubmed datasets. In addition, we achieve the best result in Macro-F1 for the Wiki-CS dataset and only lower than the ProGCL method on Micro-F1. In the Flickr dataset, we can see that the semi-supervised method GCN achieves the best performance in two metrics. Our method outperforms all self-supervised methods in the Micro-F1 metric. 

We empirically find that a two-layer GCN encoder yields the best accuracy on the Cora, Pubmed, Wiki-CS, and Flickr datasets, while a one-layer GCN encoder achieves the best performance on the Citeseer dataset. We note that some GCL methods \cite{zhu2020deep,zhang2022costa} also utilize two layers of GCN on the Citeseer dataset. This indicates that our approach achieves strong performance with fewer computations compared to the aforementioned methods on certain datasets.

\subsection{Ablation Studies}

In this section, we investigate the impact of Causal Intervention (Aug.), Invariance Objective (Inv.) and Independence Objective (Indep.) in GCIL. We design three variants of GCIL.
% (1) \textbf{GCIL\_w/o\_Inv}: GCIL without Invariance Objective (inv.) module. (2) \textbf{GCIL\_w/o\_Indep}: GCIL without Independence Objective (indep.) module. (3) \textbf{GCIL\_w/o\_Aug}: The spectral-based augmentation is removed while maintaining the optimization objective.
(1) \textbf{GCIL\_w/o\_Inv}: GCIL without Invariance Objective (inv.) module, i.e., hyper-parameter $\alpha$ and $\beta$ are set to zero, and $\gamma$ corresponds with GCIL.
(2) \textbf{GCIL\_w/o\_Indep}: GCIL without Independence Objective (indep.) module, i.e., hyper-parameter $\gamma$ is set to zero while $\alpha$ and $\beta$ corresponds with GCIL.
(3) \textbf{GCIL\_w/o\_Aug}: The spectral-based augmentation is removed while maintaining the optimization objective.
% \end{itemize}

The results of three variants on four datasets are reported in Table \ref{ablation}. From the results, we can find that the overall performance order is as follows: GCIL $>$ GCIL\_w/o\_Aug $>$ GCIL\_w/o\_Indep $>$ GCIL\_w/o\_Inv.
The invariance objective has the greatest impact on the performance, which illustrates capturing the consistency between the node representations greatly encourages the model to encode task-related information. The performance of GCIL\_w/o\_Indep is worse than GCIL, which suggests that promoting independence between dimensions encourages the model to learn more informative node representations. Additionally, the result of GCIL\_w/o\_Aug indicates that our augmentation encourages the model to learn the invariant representations better than random augmentation. Finally, GCIL obtained the best performance, which indicates the effectiveness of considering three components together.

\begin{figure}[t]
    \centering    
    \subfigure[Cora: $\alpha$] {
     \includegraphics[width=0.47\columnwidth]{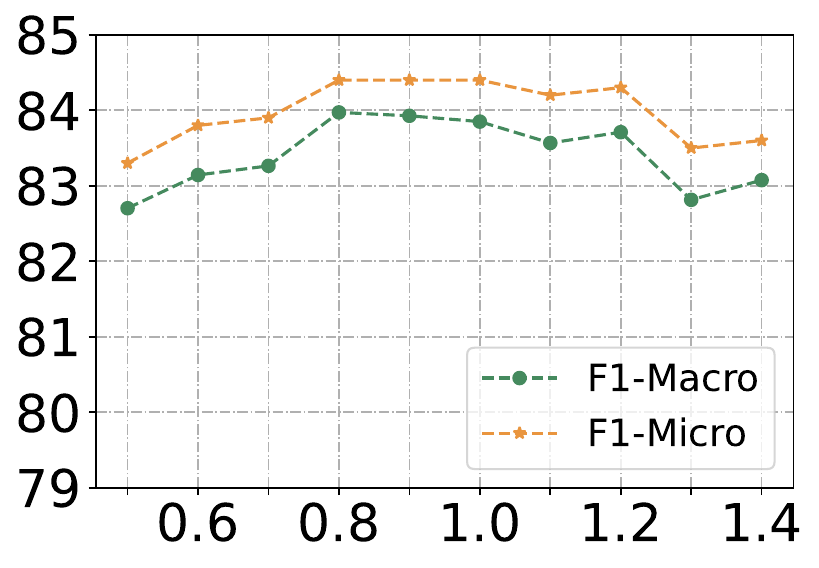}  
     \label{cora_alpha}
    }     
    \subfigure[Wiki-CS: $\alpha$] {
     \includegraphics[width=0.47\columnwidth]{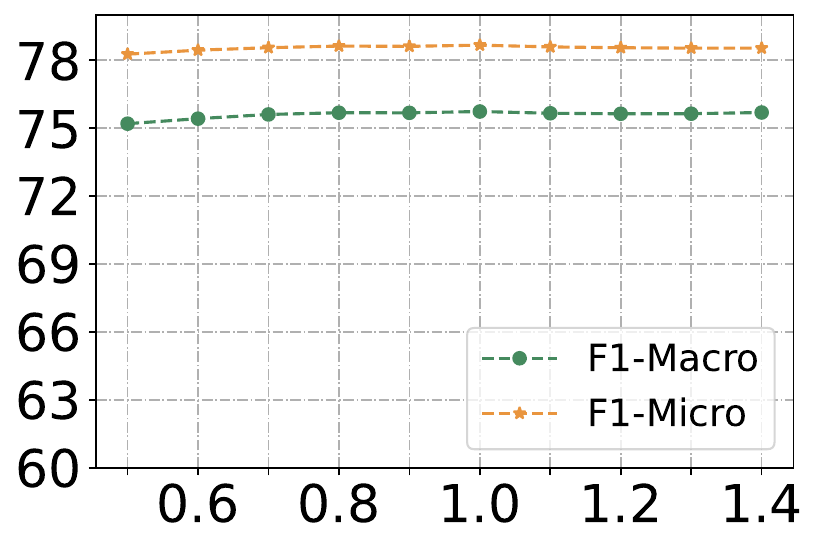}  
     \label{wiki_alpha}
    }     
    \subfigure[Cora: $\beta$] { 
   
    \includegraphics[width=0.47\columnwidth]{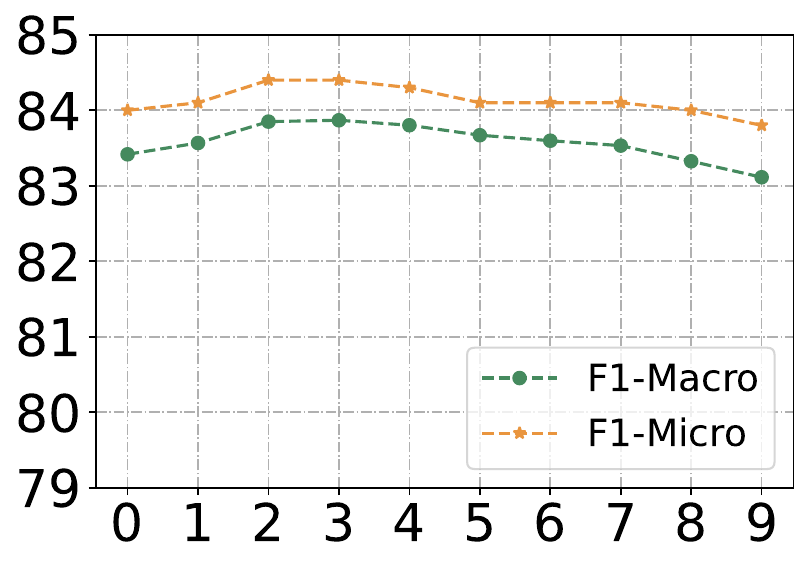}  
    \label{cora_beta}
    }    
    \subfigure[Wiki-CS: $\beta$] { 
   
    \includegraphics[width=0.47\columnwidth]{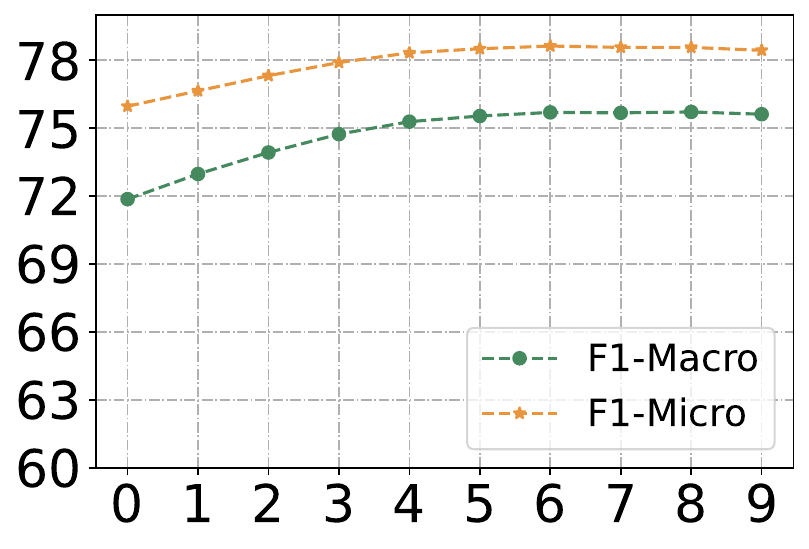}     
    \label{wiki_beta}
    }

    \subfigure[Cora: $\gamma$] {   
    \includegraphics[width=0.47\columnwidth]{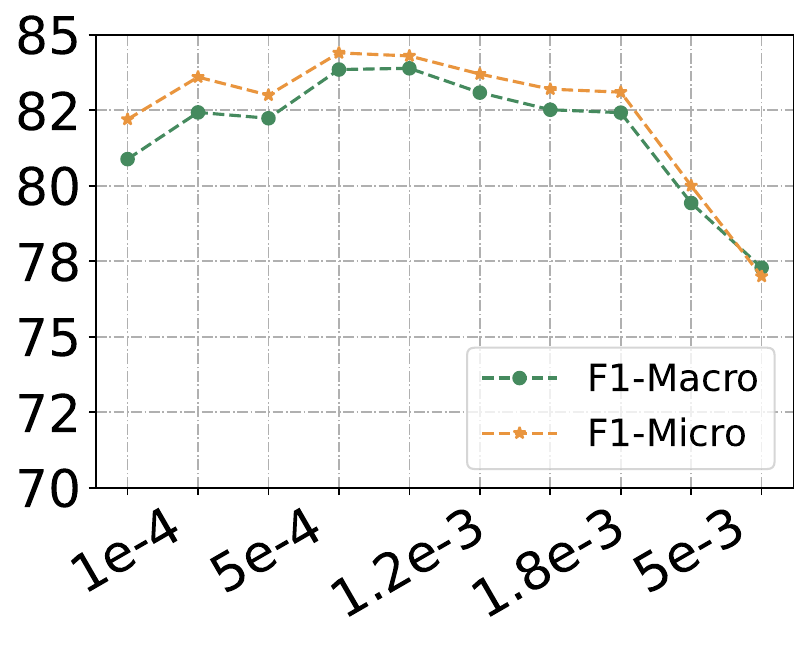}    
    \label{cora_gamma}
    }   
    \subfigure[Wiki-CS: $\gamma$] {   
    \includegraphics[width=0.47\columnwidth]{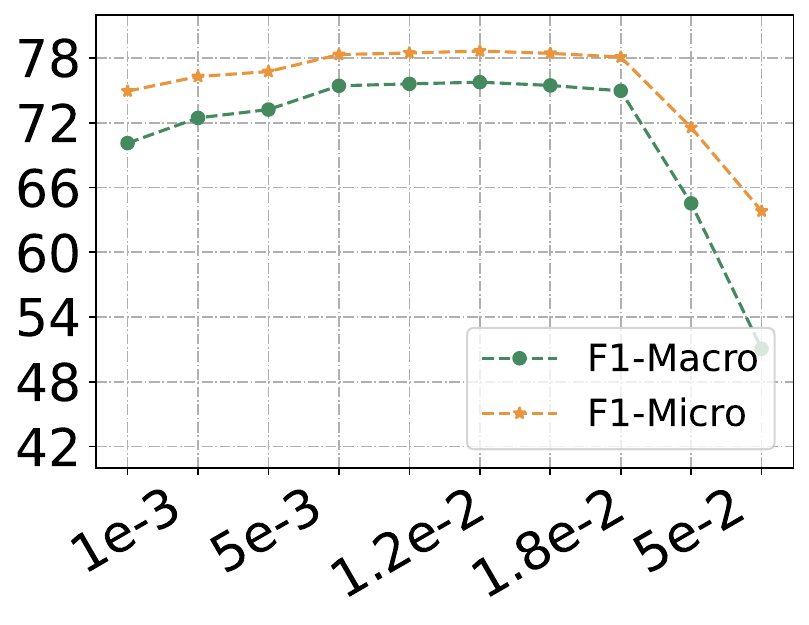}   
    \label{wiki_gamma}
    }   
    \caption{The hyper-parameter sensitivity of GCIL with varying $\alpha$, $\beta$ and $\gamma$ on Cora and Wiki-CS datasets.}     
\label{parameter_sensitity}     
\end{figure}

\subsection{Hyper-parameter Sensitivity}
In this section, we investigate the effect of $\alpha$, $\beta$, $\gamma$, and embedding dimension. The results on four hyper-parameters are respectively reported in Figure \ref{parameter_sensitity}.

\textbf{Analysis of $\alpha$.} The $\alpha$ controls the importance of the term that guarantees a constant mean of the same dimension in two representations. We fix the hyper-parameters $\beta$ and $\gamma$ while varying $\alpha$ from 0.5 to 1.4 and report the F1-Macro and F1-Micro performance of GCIL. As shown in Figure \ref{cora_alpha} and \ref{wiki_alpha}, the performance of GCIL first increases and then decreases with the growth of $\alpha$, while the performance fluctuation is very slight. It indicates that GCIL is not sensitive to $\alpha$.

\textbf{Analysis of $\beta$.}
The $\beta$ controls the importance of the term that guarantees constant standard deviation of the same dimension in two representations. Similar to the above process, we fix the hyper-parameters $\alpha$ and $\gamma$ while varying $\beta$ from 0 to 9. In Figure \ref{cora_beta} and \ref{wiki_beta}, we can see that with a suitable range, the performance of our method benefits from the increase of $\beta$, and the performance will drop slightly with the large value of $\beta$.

\textbf{Analysis of $\gamma$.}
The parameter $\gamma$ controls the contribution of the independence objective. We varied $\gamma$ from 1e-4 to 1e-2 for Cora and from 1e-3 to 0.1 for Wiki-CS to test its impact. Increasing attention on the independent term improved performance, demonstrating the objective's effectiveness in encoding relevant information. However, excessively large values of $\gamma$ sharply decreased performance, indicating poor optimization of other loss terms. We can choose the appropriate $\gamma$ by observing the degree of optimization of the invariance objective.

\begin{figure}[t!]
    \centering    
    \subfigure[w/o indep and std] {
     \includegraphics[width=0.31\columnwidth]{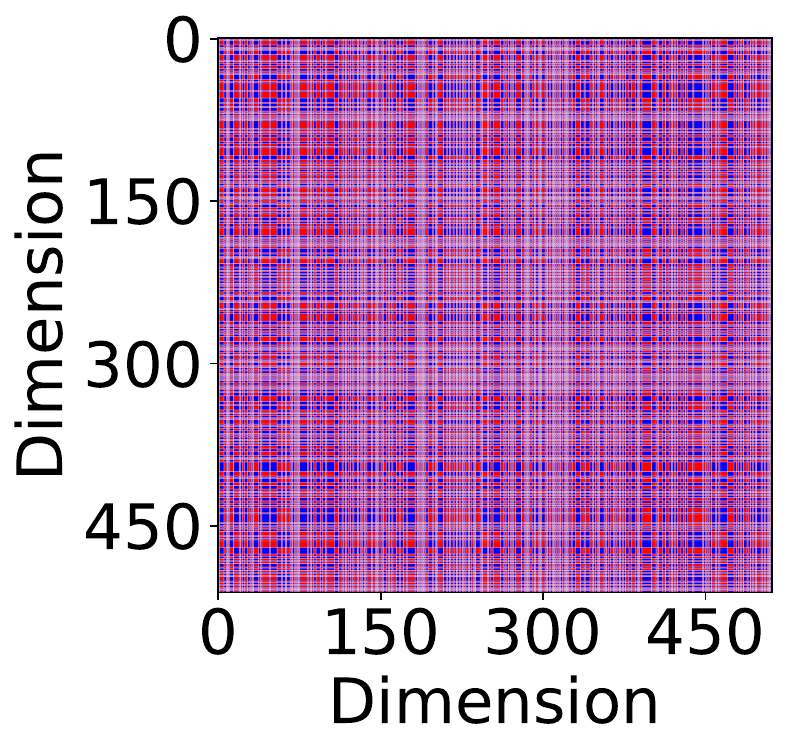}  
     \label{wiki_wo1}
    }     
    \subfigure[w/o std] { 
    \includegraphics[width=0.24\columnwidth]{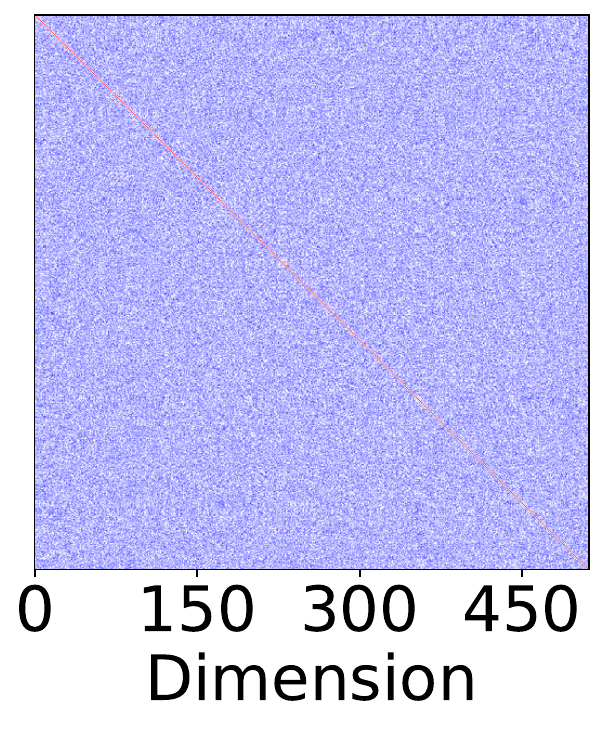}   
    \label{wiki_wo2}
    }
    \subfigure[GCIL] {
     \includegraphics[width=0.345\columnwidth]{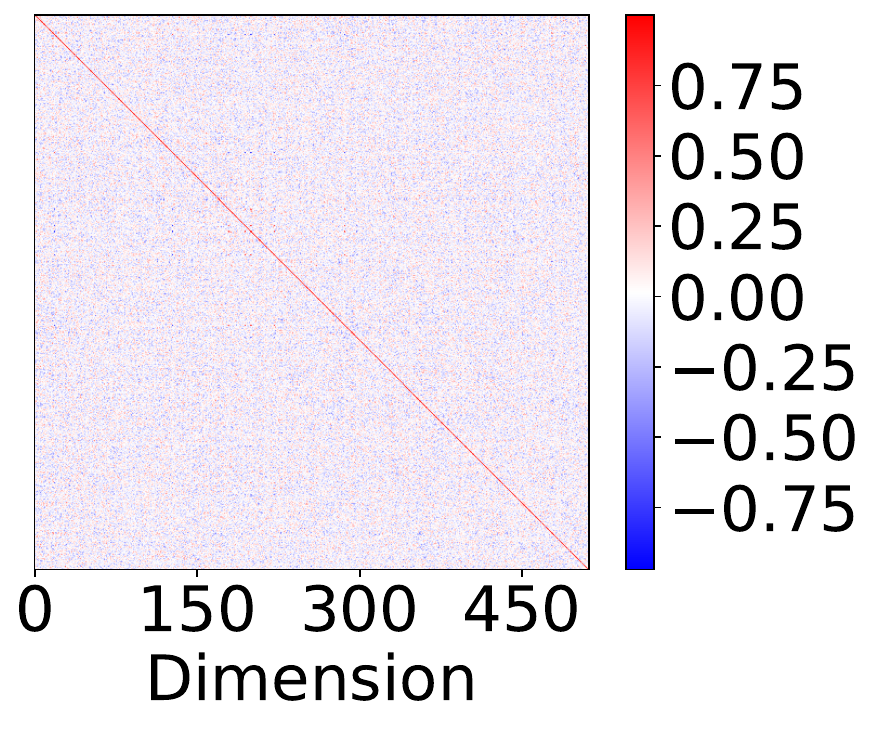}  
    \label{wiki_all}
    }
    
    \caption{Correlation matrix of the representations on Wiki-CS.}     
\label{visualization}
\end{figure}

\subsection{Visualization}
To better understand the correlated relationships between the dimensions of the representation, 
we visualize the correlation matrix of the representation matrix $Z$ for the Wiki-CS dataset. In Figure \ref{visualization}, each row and column corresponds to a representation dimension, and the color indicates the Pearson correlation coefficient \cite{cohen2009pearson}. In Figure \ref{wiki_wo1}, we set the hyper-parameter $\beta$ and $\gamma$ to 0, while keeping the other hyper-parameters unchanged. We observe that strong correlations exist among different dimensions of the representation. This suggests that these dimensions may encode similar information. In Figure \ref{wiki_wo2}, we set the hyper-parameter $\beta$ to 0 and discover that the correlations between dimensions are reduced compared to the previous case. However, different dimensions still correlate with each other. In contrast, the result of GCIL, as shown in Figure \ref{wiki_all}, the correlation matrix of the representation demonstrates that almost all off-diagonal values converge to 0. This indicates that different dimensions contain orthogonal and distinct information, which highlights the effectiveness of our method in learning informative representations.
Overall, these visualizations confirm that our proposed method successfully captures and utilizes orthogonal information in the representation, leading to improved performance in capturing meaningful features.

\section{Conclusion}
In this paper, we study graph contrastive learning from the causal perspective and find out that previous methods may discard the causal information contained in the original graph, which prevents the model from learning the invariant representations. To learn the invariant representations, we propose a novel GCL method from the causal view. We first simulate conducting intervention on non-causal factors with spectral graph augmentation. Then, we design the invariance objective and the independence objective to encourage the model to extract causal information contained in the graph. Experimental results demonstrate that our proposed GCIL obtains the best performance across baselines on four node classification datasets.

\section*{Acknowledgments}

This work is supported in part by the National Natural Science Foundation of China (No. U20B2045,
62192784, U22B2038, 62002029, 62172052, 62322203). This work is also supported by Foundation of Key Laboratory of Big Data Artificial Intelligence in Transportation(Beijing Jiaotong University), Ministry of Education (No.BATLAB202301). This project is also funded by China Postdoctoral Science Foundation (No. 2023M741946) and China Postdoctoral Researcher Program (No.GZB20230345).

% \clearpage

\bibliography{aaai24}

\end{document}